\documentclass{article}

     \usepackage[nonatbib, preprint]{neurips_2020}

\usepackage[utf8]{inputenc} % allow utf-8 input
\usepackage[T1]{fontenc}    % use 8-bit T1 fonts
\usepackage{hyperref}       % hyperlinks
\usepackage{url}            % simple URL typesetting
\usepackage{booktabs}       % professional-quality tables
\usepackage{amsfonts}       % blackboard math symbols
\usepackage{nicefrac}       % compact symbols for 1/2, etc.
\usepackage{microtype}      % microtypography
\usepackage{multirow}
\usepackage{multicol}
\usepackage{booktabs}
\usepackage[pdftex]{graphicx}

\newcommand{\mytabsep}{\\[.5em]}
\usepackage{amsmath}

\usepackage{cleveref}

\title{Automatic universal taxonomies
  for multi-domain semantic segmentation}

\author{%
  Petra Bevandić and Siniša Šegvić  \\
    Faculty of Electrical Engineering and Computing\\
  University of Zagreb\\
  Zagreb, Croatia \\
  \texttt{name.surname@fer.hr} \\
}

\begin{document}

\maketitle

\begin{abstract}
Training semantic segmentation models
on multiple datasets
has sparked a lot of recent interest 
in the computer vision community.
This interest has been motivated 
by expensive annotations 
and a desire to achieve proficiency 
across multiple visual domains.
However, established datasets have 
mutually incompatible labels
which disrupt
principled inference in the wild.
We address this issue by automatic construction
of universal taxonomies 
through iterative dataset integration.
Our method detects subset-superset relationships
between dataset-specific labels,
and supports learning of sub-class logits
by treating super-classes as partial labels. 
We present experiments on
collections of standard datasets
and demonstrate competitive 
generalization performance 
with respect to previous work.
\end{abstract}

%-------------------------------------------------------------------------
\section{Introduction}
Semantic segmentation is
an important computer vision task
with exciting applications
in intelligent transportation \cite{janai20ftcgv},
medical diagnostics \cite{nie20accv},
remote surveillance \cite{boguszewski21cvprw},
and autonomous robots \cite{hu21ral}.
Current state of the art is based on
strongly supervised learning
which induces a strong dependence
on dense semantic ground truth.
Unfortunately, producing dense annotations
requires a lot of time and money
\cite{cordts16cvpr,zlateski18cvpr}.
There are several datasets of intermediate size
\cite{neuhold17iccv,lin14eccv,zendel18eccv,zhou17cvpr},
but none that is sufficient
for delivering
robust performance in the wild
\cite{zendel18eccv}.
Thus, training across several datasets and domains
appears as an attractive research direction.
%which promises to rekindle the progress
%towards general visual recognition.

A simple baseline
involves per-dataset heads over shared features
\cite{fourure17neucom,kalluri19iccv}.
Per-dataset predictions can be recombined
into a common taxonomy \cite{zhou2021simple},
however this is not easily adapted
to multi-class problems
and overlapping taxonomies
\cite{liang2018dynamic,meletis18iv,zhao20eccv}.
%\cite{Bevandic_2022_WACV}. 
Another baseline
concatenates per-dataset taxonomies
\cite{fourure17neucom,masaki21itsc} and
feeds them to
common softmax. However, this 
may entail capacity loss
due to competition between related logits.
A recent approach reconciles
a set of taxonomies by 
pragmatic label adaptation
\cite{lambert20cvpr}
that however has to drop some classes 
in order to reduce the relabeling effort.
Recent work leverages
hand-crafted universal taxonomies
that allow superclass labels 
to promote subclass recognition 
and vice versa \cite{Bevandic_2022_WACV, liang2018dynamic, meletis18iv}.
However, this requires human judgment 
which is expensive and error prone.

\begin{figure}[thb]
\centering
    \includegraphics[width=0.95\textwidth]{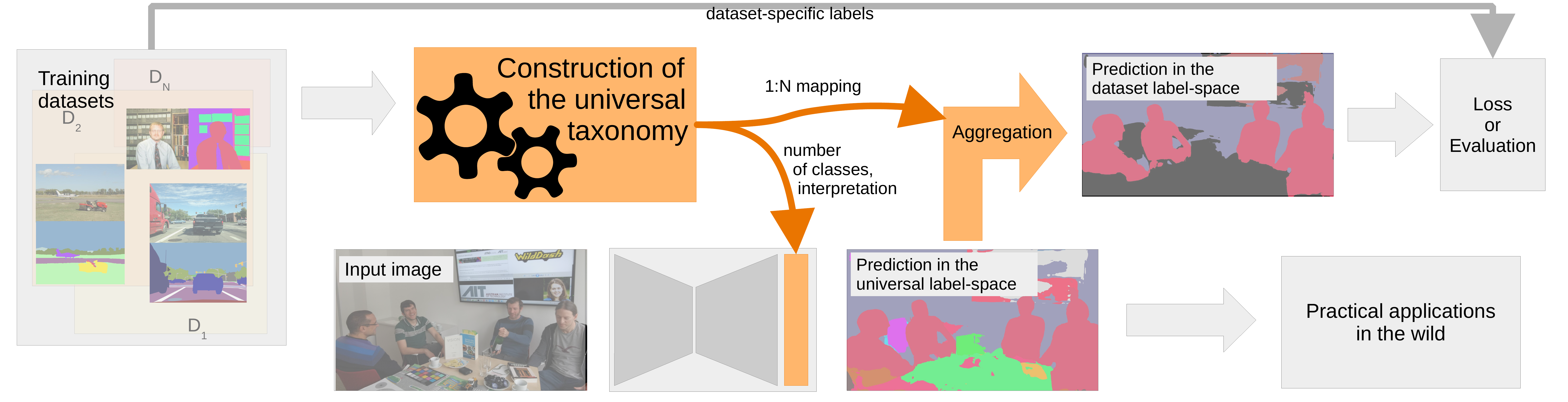}
\caption{We consider
automatic construction of a universal taxonomy
from multiple datasets with incompatible labels
(orange box).
Our method recovers a set of disjoint universal classes,
and connects each dataset-specific class
to one or more universal classes.
These 1:N mappings enable
% conversion
%of universal predictions
%into particular taxonomies,
training and evaluation on original labels
(top-right).
The universal model can be exploited
for interpretable inference in the wild
(bottom-right). 
}
\label{fig:approach-pipeline}
\end{figure}

This paper makes a step further 
by considering automatic extraction of universal taxonomies 
over incompatible datasets as sketched in Figure \ref{fig:approach-pipeline}.
Our method hypothesizes cross-dataset relations 
through co-occurence analysis.
We disambiguate these hypotheses against each other 
according to mIoU performance
on all training datasets.
We perform experiments on collections 
of large semantic segmentation datasets 
such as Vistas, Ade20k, COCO and WildDash 2. 
The recovered automatic taxonomies perform 
comparably to their manual counterparts \cite{Bevandic_2022_WACV}
while outperforming all other baselines \cite{lambert20cvpr}
by a considerable margin.

\section{Related work}

We consider semantic segmentation  
for natural scene understanding 
(sec.\ \ref{ss:rw-semseg})
by studying cross-domain models
which transcend particular training taxonomies
(sec.\ \ref{ss:rw-cross}).
We focus on automatic construction 
of universal taxonomies
(sec.\ \ref{ss:rw-univseg}).

\subsection{Semantic segmentation}
\label{ss:rw-semseg}

Semantic segmentation classifies each 
input pixel into one of C known classes
\cite{farabet13pami,shelhamer17pami}.
It is one of the most computationally intensive
computer vision tasks due to high output resolution.
The training footprint often
constrains the model capacity \cite{bulo18cvpr}.
Huge computational complexity leads
to very long training times.
Consequently, efficient models 
\cite{hong21arxiv,orsic20pr,nie20accv}
and knowledge transfer
\cite{he19iccv}
are a good fit 
for large cross-domain experiments.
Besides faster training,
they also improve accessibility 
and decrease environmental impact
\cite{schwartz20cacm}.

\subsection{Cross-domain training}
\label{ss:rw-cross}

Early  
cross-domain training approaches
%do not attempt to digest individual taxonomies
%towards some universal taxonomy 
do not incorporate
relations between 
individual taxonomies.
%with improved properties.
Instead, they either use separate dataset-specific
prediction heads on top of shared features
\cite{kalluri2019universal},
or train on a concatenation
of particular taxonomies \cite{masaki21itsc}.
Naive concatenation has been improved
by encouraging cross-talk between logits
\cite{fourure17neucom}.

Training dense open-set recognition models
on positive and negative data 
%\cite{bevandic19gcpr,chan21iccv,biase21cvpr}.
may improve detection of unknown
\cite{chan21iccv,biase21cvpr}
or novel classes \cite{uhlemeyer22arxiv}.
%\cite{bevandic18arxiv,}.
This can be viewed
as asymmetrical cross-domain training.
The positive domain corresponds to
the primary recognition task
%which specialize for a particular domain 
(e.g.\ road driving) while 
the negative domain typically corresponds 
to anomalies \cite{chan21iccv,BEVANDIC22ivc, tian2021arxiv}.

Some cross-domain approaches
propose hierarchical universal taxonomies
with distinct nodes for categories and classes 
\cite{liang2018dynamic,meletis18iv}.
However, this requires complex learning
procedures while not offering advantages
over flat universal taxonomies.

Incompatible datasets can be unified
under a custom common taxonomy
by manual relabeling and removal of
subclasses \cite{lambert20cvpr, Zendel_2022_CVPR}.
However, these modifications are
tedious and destructive.
The more datasets one converts, the harder it gets 
to extend the common taxonomy with new subclasses.
%They perform an extensive manual relabeling effort
%in order to consistently label these 194 classes
%across a collection of 7 datasets denoted as MSeg.
%However, their approach has two distinct shortcomings.
%First, they had to omit 61 dataset-specific % or 100?
%classes in order to contain the relabeling effort.
%Furthermore, those taxonomies cannot be extended
%without triggering a costly round of relabeling
%thousands of images in the existing dataset.
%These shortcomings suggest that it may be a good idea
%to embrace subset/superset relationships 
%instead of avoiding them.
%A key capability at this point is being able 
%to learn from labels with different granularity.
%If we could do that, we could manually construct
%a universal taxonomy for a given set of datasets,
%and proceed with cross-domain training 
%without having to relabel a single pixel.
This issue can be elegantly solved 
by constructing a universal taxonomy 
where each dataset-specific class 
can be expressed as a union of universal classes 
\cite{Bevandic_2022_WACV}.
In this case,  universal logits can be trained
with respect to dataset-specific labels
(cf. Figure \ref{fig:approach-pipeline})
since dataset posteriors correspond to 
sums of universal posteriors \cite{cour2011learning}. 
The result of such construction
allows principled cross-dataset training 
without any modification of the original datasets.
We extend this approach 
by considering automatic construction 
of such universal taxonomy 
from datasets with differing granularities.

\subsection{Automatic construction of universal taxonomies}
\label{ss:rw-univseg}

%Previous discussion shows that
%manual universal taxonomies
%go a fairly long way towards
%seamless cross-domain training.
%However, there are 
%at least three reasons why 
%it would be beneficial
%to automate this process.
%First, we can never rule out a human error
%within the code for universal taxonomy.
%Such bug would decrease the performance 
%of derived models for only a few mIoU points
%and go unnoticed for years.
%Second, manual taxonomies are
%typically assembled 
%by reasoning about the label semantics
%instead by looking at the datasets.
%However, label semantics may be ambiguous.
%For instance, the visual label tomato 
%may denote plant, fruit, salad
%or processed food.
%Finally, automatic taxonomies
%have a much larger scalability 
%potential than manual taxonomies.
%It is clear that the task of
%merging hundreds of datasets 
%with thousands of classes
%is much better suited for computers
%instead for humans.
%These three problems increase each other:
%human errors and label ambiguity
%become much more likely 
%as datasets get larger.
Manual resolution
of dataset discrepancies is error prone,
especially
when the ambition is to train on 
multiple large-scale
datasets with hundreds of symbolic labels.
%This is 
%compounded by the fact
%%labels may be ambiguous and context dependent,
%which may mean that relying on the class names
%may not be enough.
This issue can be elegantly
curcumvented by expressing
semantic labels with
text embeddings 
instead of categorical distributions 
\cite{li22iclr,yin22arxiv}.
However, a recent study
reveals that label semantics
often vary across datasets. 
Their experiments suggest that visual 
cues outperform label
semantics as a tool for recovering
cross-dataset relations \cite{uijlings2022eccv}.

Recent work constructs an automatic taxonomy
for object detection \cite{zhou2021simple}.
Their approach starts by training 
a model with shared features 
and separate prediction heads
\cite{fourure17neucom,kalluri19iccv}.
Subsequently, they freeze the trained features
and optimize dataset-specific
mappings through linear programming.
%This model is used
%to automatically
%identify identical
%classes in different datasets using
%integer linear programming.
The resulting cross-dataset mappings
outperform their text-embedding
counterparts.
However, this approach
does not handle subset/superset
relationships and therefore
does not produce a true taxonomy
when dataset-specific classes happen to overlap.
This situation hampers multi-class
performance due to competition
between related logits \cite{Bevandic_2022_WACV}.
%Partially merged taxonomies are
%inferior to full universal taxonomies
%because they retain competition 
%between overlapping classes.

Cross-dataset relations have also been recovered
according to class names \cite{lambert20cvpr}.
In this setup, superclass logits can be trained
with subclass labels \cite{kim22arxiv}.
However, this setup can not accommodate the
standard multi-class loss,
fails if there is a name mismatch \cite{uijlings2022eccv, zhou2021simple}, 
and cannot train subclass logits 
with superclass labels. 

Different than all previous work, 
our method constructs the only flat universal taxonomy 
which retains all labels
in presence of subclass/superclass relations.

\section{Method}
\begin{figure}[htb]
\centering
\includegraphics[width=0.9\textwidth]
  {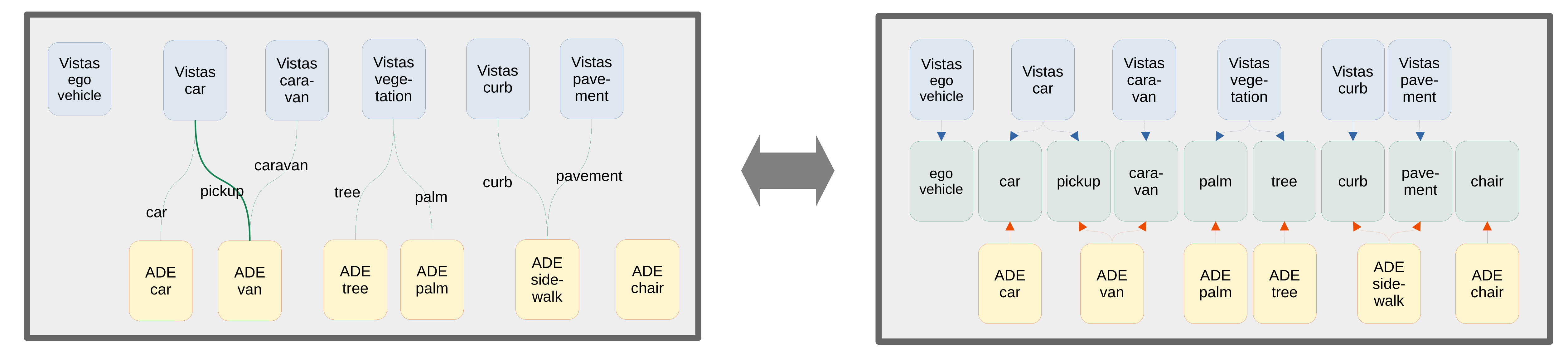}
\caption{
We collect information about cross-dataset relations
and expres it as a bipartite graph 
that connects classes of the two datasets (left).
We analyze the bipartite graph in order to 
recover the universal taxonomy (right).
Each edge of the bipartite graph identifies
the visual concept shared by the related classes.
}
\label{fig:univ-to-overlap}
\end{figure}

We consider automatic recovery of a flat
universal taxonomy for a given collection
of datasets in order to allow 
cross-domain training of dense
prediction models.
%Our universal models can predict classes
%which are not in the evaluation taxonomy
%which we dub foreign or extra-domain predictions.
%Conversely, predictions within the evaluation 
%taxonomy are called intra-domain predictions.
%The procedure for extracting manual universal taxonomies
%through iterative modification of label collections
%\cite{Bevandic_2022_WACV} is difficult to automate.
We propose to automatically discover 
hierarchical relations between classes 
of the two datasets
and use this information
to construct the universal taxonomy
as illustrated in 
Figure \ref{fig:univ-to-overlap}.
%We construct pairwise taxonomies
%by hypothesizing and testing relations
%between pairs of classes from different datasets.
%We show that individually trained models [33]
%are worse than naive concatenation models for this task
%since the latter produce better co-occurrence information
%and can disambiguate competing hypotheses.
We extend pairwise taxonomies
for arbitrary tuples of datasets
through tournament-style iteration.

Occasionally, our cross-domain models will predict classes
which are disjoint from the native taxonomy of the input image.
We denote such occurrences as foreign
or extra-domain predictions.
Conversely, predictions which fall within the native taxonomy
are denoted as intra-domain predictions.

\subsection{Universal taxonomy for two datasets}
\label{ss:merge-pair}
Let us consider two dataset-specific taxonomies 
as $T_a=\{c^a_i\}$ and $T_b=\{c^b_j\}$.
We apply a model trained for $T_b$
to training data with $T_a$ groundtruth
and the other way round.
We collect co-occurrence statistics
between ground-truth classes and foreign predictions
and store them into two co-occurrence matrices
$|T_a|\times|T_b|$ and $|T_b|\times|T_a|$.
%where rows represent the ground-truth 
%and columns the foreign prediction.
For convenience, we shall denote 
the most common foreign prediction
for a ground-truth class $c$ as mcfp($c$).
We shall hypothesize relations between datasets
by considering a bipartite graph
induced by the mcfp function.
The graph has $|T_a|$ + $|T_b|$ vertices 
which represent classes,
and $|T_a|$ + $|T_b|$ edges 
pointing from a ground-truth class
to its most frequent foreign prediction.
Hence, each vertex
has exactly one outgoing edge.
This choice increases the statistical power of our hypotheses
and reduces the number of hypotheses and hyper-parameters.

We illustrate our approach 
on the following two taxonomies: 
ADE20K = \{’ade-road’,...\} and 
Vistas = \{’vistas-road’, ’vistas-zebra’,...\}. 
The class 'ade-road' is a superset of 
'vistas-road' and 'vistas-zebra'.
In Vistas images we shall typically have
mcfp(’vistas-road’) = mcfp(’vistas-zebra’) = ’ade-road’. 
On the other hand, in ADE20k images we will have:
mcfp(’ade-road’) = ’vistas-road’.
We observe that the mcfp statistic
suffices to hypothesize that 'ade-road' 
is a superset of 'vistas-road' and 'vistas-zebra'. 

\begin{figure}[htb]
\centering
\includegraphics[width=0.98\textwidth]
{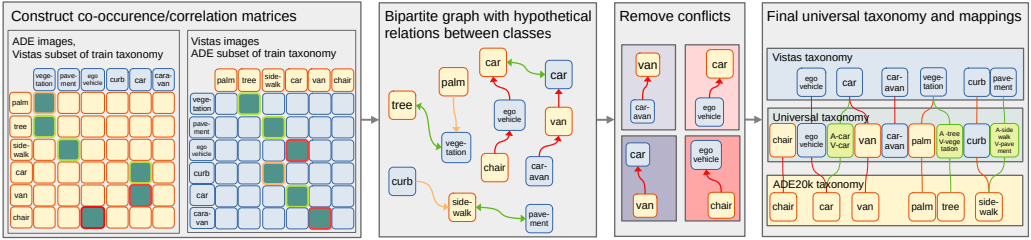}
\caption{Our method
  %We perform inference by means of 
  %the naive concatenation model
  collects two co-occurrence matrices (left)
  between the ground truth (rows) and 
  foreign predictions (columns). 
  For each row, we locate the strongest column
  and form a bipartite graph (center-left). We note
  three 2-cycles (green),
  two asymmetric relationships (orange),
  and two inconsistent triplets (red).
  We form pairs of
  conflicting hypotheses (shades of the same color,
  center right) and
  resolve them by comparing performance on the training dataset
  (cf. Figure \ref{fig:conflict-resolution}).
  The collected evidence allows us to
  recover the universal taxonomy (right).
  %(cf. Figure \ref{fig:method2}).
}
\label{fig:method}
\end{figure}

We analyze the bipartite graph as follows.
Cycles of length 2
($c^a_i \rightarrow c^b_j \rightarrow c^a_i$)
indicate overlap. 
Asymmetric relationships
($c^a_i\rightarrow c^b_j$)
suggest a subset hypothesis $c^b_j \supset c^a_i$.
Inconsistent triplets
$c^a_i \rightarrow c^b_j \rightarrow c^a_k$
where $c^a_k \not\rightarrow c^b_j$
suggest a subset and a superset hypothesis 
$c^b_j \supset c^a_i \land c^a_k \supset c^b_j$.
This would mean that $c^a_i \cap c^a_k \neq\emptyset$,
which is impossible since input datasets
have proper taxonomies.
We consider
$c^a_i \subset c^b_j$ and $c^b_j \subset c^a_k$
as competing hypotheses 
which we disambiguate in \ref{ss:naive}.
Figure \ref{fig:method}
illustrates this procedure 
on the ADE20K-Vistas example.

We recover the final universal taxonomy 
from the disambiguated bipartite graph 
(cf. Figure \ref{fig:conflict-resolution}, right).
The graph associates each universal class 
with all incident dataset-specific classes.
We thus base the names of universal classes 
on associated dataset-specific classes: 
a one-way edge inherits the name of its source vertex,
while a two-way edge inherits the names 
of both adjacent vertices.
If we have ade-car$\rightarrow$ vistas-car 
and vistas-car $\rightarrow$ ade-car,
we would have a universal class named 'ade-car/vistas-car'.
Such naming convention 
provides a degree of interpretability
to resulting universal models.

\subsection{Conflict resolution with improved naive concatenation}
\label{ss:naive}

Naive concatenation is a 
cross-domain pseudo-taxonomy
which we obtain by aggregating 
dataset-specific taxonomies.
We use the term pseudo-taxonomy
since its members may overlap
(e.g. ade-road and vistas-road).
Such overlaps require discrimination 
of semantically related concpets
and consequently diminish 
the effective model capacity.

Naive concatenation performance can be improved
by performing post-inference mapping 
towards the evaluation taxonomy.
We define unnormalized classification score 
of a particular evaluation class $S(c^a_i)$
as the sum of its posterior $P(c^a_i)$
with posteriors of intersecting foreign classes $P(c^b_j)$ 
\cite{Bevandic_2022_WACV}:

\begin{equation}
  S(c^a_i) = P(c^a_i) + \sum_{c^a_i \cap c^b_j \neq \emptyset } P(c^b_j)
  \label{eq:evalmap}
\end{equation}

The expression $c^a_i\cap c^b_j \neq \emptyset$ is true
when $c^a_i$ and $c^b_j$ are in any kind of relation.
The model prediction corresponds to 
$\operatorname*{argmax}_{i} S(c^a_i)$.

We illustrate the recovery of different dataset-specific scores
over ADE20K and Vistas as follows:
$S(\mbox{ade-road}) = 
  P(\mbox{ade-road}) + P(\mbox{vistas-road}) + 
  P(\mbox{vistas-zebra})$,
  $S(\mbox{vistas-road}) = 
  P(\mbox{vistas-road}) + 
  P(\mbox{ade-road})$
  and
$S(\mbox{vistas-zebra}) = 
  P(\mbox{vistas-zebra}) + 
  P(\mbox{ade-road})$.
  
\begin{figure}[htb]
  \centering
  \includegraphics[width=0.98\textwidth]
  {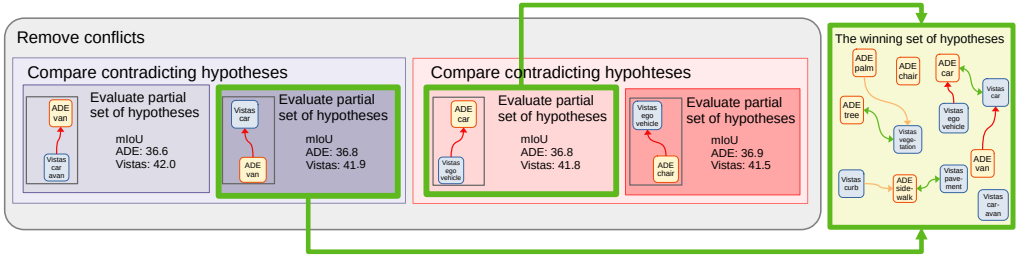}
    
  \caption{Resolution of contradicting hypotheses 
    according to the train performance
    of the naive concatenation model
    with post-inference mapping.
    The winning hypotheses are selected according to 
    the unnormalized classification score (\ref{eq:evalmap})
    on all train datasets.
    We resolve one pair of hypotheses at a time. 
    This requires $2 N_C N_D$ evaluations 
    where $N_C$ is the number of conflicting pairs 
    and $N_D$ indicates the number of training datasets. 
  }
  \label{fig:conflict-resolution}
\end{figure}

We can use post-inference mapping
to compare competing hypotheses 
arising from inconsistent triplets 
(\S\ref{ss:merge-pair}).
We create a post-inference mapping 
for each competing hypothesis 
and evaluate performance
according to (\ref{eq:evalmap}).
We choose the hypothesis with
the highest train mIoU performance
averaged over all involved datasets.
The resolution involves
$2 N_D N_C$ automatic evaluations 
where $N_C$ denotes 
the number of conflicting pairs and 
$N_D$ indicates the number of training datasets.
This procedure is illustrated in Figure
\ref{fig:conflict-resolution}.

\newcommand{\myheight}{4cm}

\subsection{Universal taxonomy for more than two datasets}

Recovering a pairwise universal taxonomy
allows us to unify the two involved datasets.
The resulting meta-dataset
contains images from the two datasets
and partial labels in form of 
unions of universal classes.
We can proceed by unifying this meta-dataset 
with subsequent datasets.
However, lack of proper ground-truth precludes recovery 
of the proposed co-occurrence matrix.
We therefore approximate the co-occurrence matrix 
with a co-incidence matrix 
between intra-domain and foreign predictions.

To recover a universal taxonomy 
for more than two datasets,
we proceed iteratively.
We start by forming
pairwise universal taxonomies.
We then train naive concatenation models
over pairs of meta-datasets
and use them to unify the involved meta-datasets.
We formulate mapping 
functions for original datasets
as compositions of intermediate
mapping functions.
%by chaining them through 
%the intermediate taxonomies
%towards the final taxonomy.

The proposed procedure can be applied
to any number of datasets
in a straight-forward manner.
We have successfully applied this procedure
in order to recover the universal taxonomy 
for the MSeg dataset collection.

\section{Experiments}
We train semantic segmentation models
in multi-domain setups.
We promote efficient experimentation \cite{schwartz20cacm}
by leveraging pyramidal SwiftNet \cite{orsic20pr}
with three shared ResNet-18 \cite{he16cvpr}
backbones and ImageNet pre-training (SNp-rn18).
We train on automatic universal taxonomies
with partial labels
\cite{cour2011learning,zhao20eccv,Bevandic_2022_WACV}. 
We train naive concatenation models
with the standard NLL loss and
the multi-head model with
a sum of head-specific NLL losses. 
Both losses prioritize
pixels at
semantic boundaries \cite{zhen19aaai}.
We perform early stopping
with respect to average mIoU 
validation performance.
We attenuate the learning rate
between $5\cdot10^{-4}$ and $6\cdot10^{-6}$
through cosine annealing.
We evaluate by mapping foreign
predictions to the void class  \cite{cordts16cvpr, Bevandic_2022_WACV}.
%Void predictions increase the count
%of false negative predictions
%for the given ground truth class.

We train on random crops of
512$\times$512 (\S\ref{ss:multi-results} and \S\ref{ss:mseg-results}) or
768$\times$768 pixels
(\S\ref{ss:pairs-result})
with horizontal
flipping and random scaling between
0.5$\times$ and 2$\times$.
We favour crops with rare classes
and form batches with even representation
of all datasets.
Our universal models were trained on one
Tesla V100 32GB. We train naive
concatenation models on two GPUs in
order to ensure the same batch
size across considered dataset collections.
We construct universal taxonomies by
analyzing only the training subsets.

\subsection{Unifying dataset pairs}
\label{ss:pairs-result}

We present experiments on pairs of datasets
with incompatible taxonomies.
We compare our automatic universal taxonomy
with two baselines as well as with a
manually constructed  universal taxonomy
\cite{Bevandic_2022_WACV}.
The two baselines are the naive taxonomy
and a model with per-dataset prediction heads
and a separate dataset detection head.
All models are trained for 100 epochs.
%we use the full
%training dataset to construct the universal 
%taxonomy.
%We compare the performance 
%of our automatic taxonomies
%to naive baselines and our manual
%universal taxonomies.

Table \ref{table:pairs-conf-wd-mvd} presents results of unifying
Vistas \cite{neuhold17iccv} (road-driving,
65 classes) and WilDash 2 
(WD2, road-driving, 25 classes) \cite{zendel18eccv}.
We split WD2 into minitrain and minival as 
in \cite{Bevandic_2022_WACV}.
Vistas has a finer
granularity than WD2
with the exception of car types.
%Vistas is a road-driving datasets 
%with 65 classes, while
%ADE20K is a general dataset with 150 classes.
%ADE20K also recognizes
%some road-driving classes, but has a coarser
%granularity than Vistas.
%which we form
%by using the first 572 images for validation
%and the remainder of the dataset for
%training.
%In the WD2-Vistas experiment we also
%compare the two
%conflict resolution approaches, 
%but only use the pairwise method
%for ADE20k-Vistas experiment 
%due to exponential complexity 
%in the number of class pairs.
%These experiments are shown
%in 
Our automatic universal taxonomy performs comparably
to the manual universal taxonomy
while outperforming the multi-head baseline 
as well as naive concatenation.%baseline. 
Interestingly, our automatic
taxonomy outperforms the manual taxonomy
on some rare Vistas classes,
which is likely due to their association
with more frequent WD2 classes
(e.g.\ wd-person and vistas-ground-animal,
and wd-truck and vistas-trailer).

\begin{table}[htb]
  \centering
  \begin{tabular}{lccccc}
    Taxonomy & \#  & evals & WD2   & Vistas
    \\
    \toprule
    two heads + dataset recognition & 65 + 33 + 2 & N/A & 54.0 & 42.2
    \\
    naive concat  & 98 & N/A & 54.8  & 42.8
    \\
    manual univ. & 67 & N/A & 56.2 & 44.4
    \\
    \midrule
    %auto univ. (conf.) & 68 & 4 & 54.4 & 44.3 \\
    %auto univ. (con, full) & N/A & N/A & N/A & N/A && 67 & 8 & 54.9 & 44.7\\
    auto univ. (ours) & 67 & 4 & 54.6 & 45.9 \\
    %auto univ. (coinc, full) & N/A & N/A & N/A & N/A && 67 & 8 & 54.8 & 45.3
    \mytabsep
\end{tabular}
\caption{Evaluation of joint training on WD2 and Vistas.
Columns show the number of logits (\#),
number of tested
hypotheses (evals) and mIoU performance
on both datasets.}
\label{table:pairs-conf-wd-mvd}
\end{table}

Table \ref{table:pairs-conf-ade-mvd} pairs Vistas with
ADE20K \cite{zhou17cvpr} 
(photos, 150 classes). 
This experiment also validates
two approaches for collecting evidence
about visual similarity of dataset-specific classes. 
We compare separately trained per-dataset models \cite{uijlings2022eccv} 
with the naive concatenation baseline.
We also validate the two conflict resolution approaches
based on co-occurrence and co-incidence matrices. 
Note also that conflict resolution 
is not feasible with separate per-dataset models. 

We observe that the universal taxonomy
produced by separate models
has almost as many logits as the
naive concatenation baseline.
Our automatic universal taxonomy for
ADE20k-Vistas has less training logits
than its manual counterpart.
Interestingly, it hypothesizes less relations 
than the manual approach (182 < 186)
even before the contradicting
hypotheses are resolved.
This happens because our automatic
method connects some classes
that are kept separate in the manual taxonomy 
(e.g. connecting flags with banners or 
 rail tracks with conveyor belts). 
Coincidence matrices perform similarly
to co-occurrence matrices,
although their universal taxonomies differ.

%and, with the exception
%of WD 2, match
%the manual universal taxonomy.
%The performance drop on WD 2
%likely happens due to noisy mappings,
%(e.g. WD-trailer  maps to
%\{Vistas-trash-can\}).%, while WD-truck maps to
%\{WD-truck/Vistas-truck, Vistas-trailer and
%Vistas-other-vehicle\}.
%Learning with partial labels favours the winning logit 
%during training and is able to tolerate
%some noise.
%This happens because our approach sometimes
%makes connections based on visual similarity.
%Smaller number of training logits in the
%ADE20k-Vistas experiments reflects
%a few incorrectly merged
%classes (e.g. vistas-fire hydrant and 
%ade-vase)

\begin{table}[htb]
  \centering
  \begin{tabular}{lccccccccc}
    Taxonomy  & \#  & evals & ADE  & Vistas
    \\
    \toprule
    naive concat  &215 & N/A & 36.8  & 41.1
    \\
    manual univ. & 186 & N/A & 37.4 & 42.7 
    \\
    \midrule
    auto univ. (separate models, co-occurrence) &  213 & N/A & 37.4 & 41.7 \\
    auto univ. (concat, co-occurrence) & 178 & 24 & 37.4 & 42.8 \\
    %auto univ. (con, full) & N/A & N/A & N/A & N/A && 67 & 8 & 54.9 & 44.7\\
    auto univ. (concat, coincidence) & 176 & 26 & 36.9 & 42.5 \\
    %auto univ. (coinc, full) & N/A & N/A & N/A & N/A && 67 & 8 & 54.8 & 45.3
    \mytabsep
\end{tabular}
\caption{Evaluation of joint training on ADE20K-Vistas.
Columns show number of logits (\#), 
number of tested hypotheses (evals) 
and mIoU performance on both datasets.
Automatic construction of universal taxonomy
with separate per-dataset models
underperforms with respect to 
the taxonomies built with naive concatenation.
Collecting evidence through
co-occurrence and coincidence
performs comparably.}
\label{table:pairs-conf-ade-mvd}
\end{table}

\subsection{Merging multiple datasets}
\label{ss:multi-results}
Table
\ref{table:threedatasets} 
evaluates our universal taxonomy over three datasets.
We start from the universal taxonomy
ADE20K-Vistas
and extend it through unification with
COCO (photos, 133 classes) \cite{lin14eccv}.
Due to the huge size of the COCO dataset,
we decrease the number of training epochs to 20.
Before cropping, each image is resized so that its smaller
side is 1080 pixels. 
We train our models on full training datasets, but only use
the first 10000 images from each dataset
for automatic construction of the universal taxonomy.
The table shows that automatic universal
taxonomy outperforms
naive concatenation
while substantially reducing the number
of training classes.

\begin{table}[htb]
  \centering
  \begin{tabular}{lcccccc}
    Taxonomy && \# & evals &  ADE  & Vistas & COCO
    \\
    \toprule
    naive concatenation   && 348 & N/A & 30.7 & 32.7 & 36.5
    \\
    manual univ.  && 243 & N/A & 31.3 & 39.0 & 34.6
    \\
    \midrule
    auto univ. && 233 & 44 & 30.8 & 37.4 & 37.7 \\
  \mytabsep
 \end{tabular}

\caption{Joint training on
ADE20K, Vistas and COCO.
Columns show
the number of logits (\#),
number tested hypotheses (evals)
and mIoU performance.
}
\label{table:threedatasets}
\end{table}

\begin{figure}[htb]
    \includegraphics[width=0.98\textwidth]{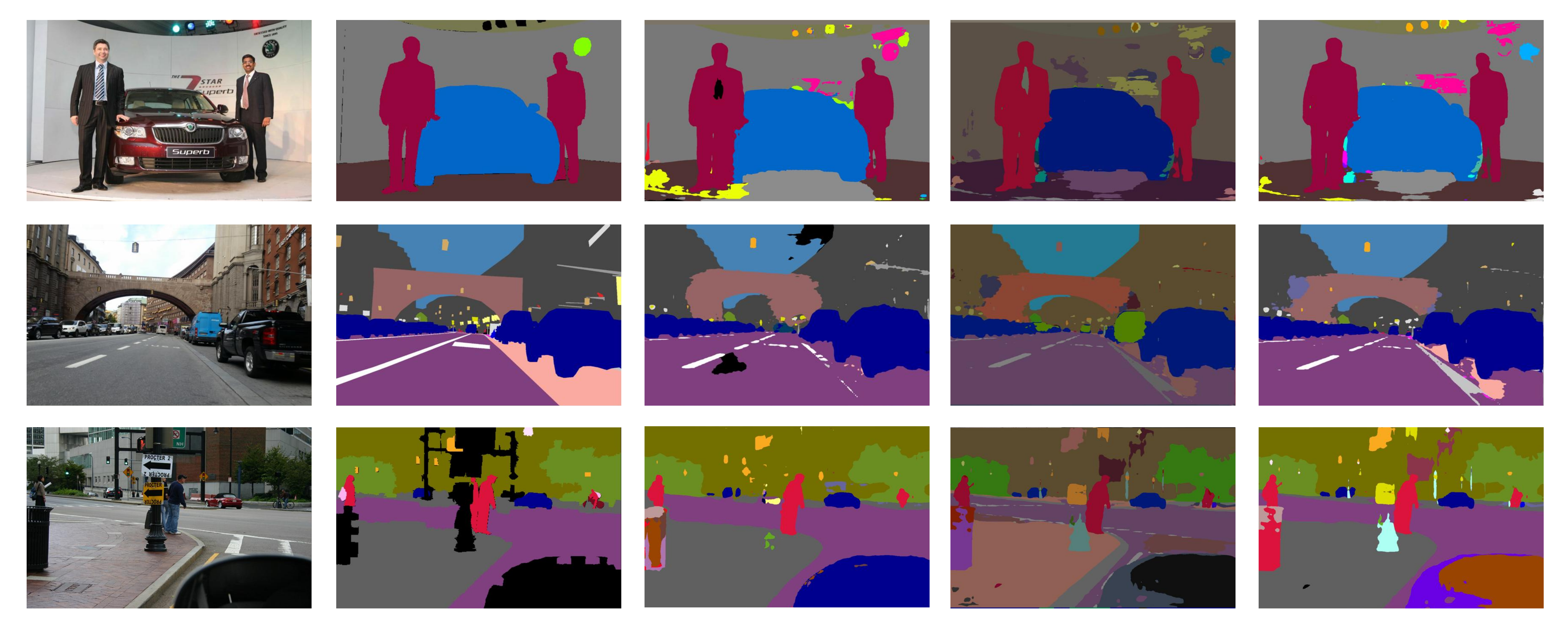}
\caption{Qualitative comparison
of cross-domain models
on ADE20K (top), Vistas (middle)
and COCO (bottom).
We show the input image (column 1),
ground truth labels (column 2),
predictions of the naive
concatenation model (column 3),
and predictions of our model
in universal (column 4)
and dataset-specific
labels (column 5). 
Naive concatenation introduces competition
between logits that represent
the same visual category.
This triggers void predictions (black)
on sky and road in the Vistas image.
Our universal model finds
universal classes
that are not present in the
corresponding dataset-specific taxonomy
and connects them with correct
dataset-specific classes:
road-marking $\rightarrow$ road
and curb $\rightarrow$ sidewalk in COCO,
tie $\rightarrow$ person in ADE20K
and van $\rightarrow$ car in Vistas. 
}
\label{fig:images}
\end{figure}

Figure \ref{fig:images}
presents a qualitative comparison
between our automatic taxonomy and
naive concatenation.
Our automatic taxonomy succeeds to actualize 
many good class connections,
such as mapping ade-food to \{ade-food/coco-donut, coco-pizza,
coco-sandwich, coco-hot-dog, coco-carrot, coco-food-other\}.
%or coco-mountain-merged to \{ade-hill, ade-land,
%vistas-mountain/ade-mountain/coco-mountain\}.
Furthermore, it finds some coherent connections
we did not initially consider
in our manual taxonomy
such as mapping
'ade-person' to \{'vistas-bicyclist',
  'vistas-person/ade-person/coco-person',
  'coco-baseball glove',
  'coco-tie'\}.
 
\subsection{Large-scale experiment on the MSeg dataset collection}
\label{ss:mseg-results}
The MSeg dataset collection \cite{lambert20cvpr}
encompasses
ADE20K \cite{zhou17cvpr},
BDD (19 classes) \cite{yu18bdd}, 
Cityscapes (28 classes) \cite{cordts16cvpr}, 
COCO \cite{lin14eccv}, 
IDD (31 classes) \cite{varma19wacv},
SUN RGBD (37 classes) \cite{song15cvpr}
and Vistas  \cite{neuhold17iccv}.
The authors of the MSeg collection
adapt all seven datasets towards a custom universal
taxonomy of 194 classes.
However, their taxonomy entails
an omission of 61
classes in order to contain 
the relabeling effort.
Note also that adding a new class
to the MSeg taxonomy would require
manual relabelling of all seven datasets.

We start the recovery by unifying dataset pairs: BDD-Cityscapes, 
IDD-Vistas, and ADE-COCO. We
proceed by unifying BDD-City with IDD-Vistas, and ADE-COCO with
SUN RGBD. Finally, we construct the universal 
taxonomy over all 7 datasets. 
If COCO is among
the training datasets, we train for 20 instead of 100 epochs.

Table \ref{table:univ-mseg}
compares our automatic universal taxonomy
to the manual universal taxonomy
and the MSeg taxonomies.
Our automatic taxonomy performs comparably
to the manual universal taxonomy \cite{Bevandic_2022_WACV}
while outperforming MSeg taxonomy and naive concatenation.
Our automatic taxonomy
contains less classes than the
manual universal taxonomy.
This happens due to a few incorrect 
associations between rare classes 
such as equating city-caravan, ade-washer and coco-toaster.
Furthermore, our approach brings 
some debatable but arguably correct decisions 
due to visual similarity. 
For example, vistas-pole is associated with
bdd-pole/city-pole/vistas-pole/ade-pole,
coco-baseball bat, coco-skis, sun-night-stand,
and ade-column-pillar.

Interestingly, our approach finds
some potentially valid connections 
that we did not initially consider in our manual taxonomy.
For instance, it associates vistas-water to ade-swimming pool
(there is often water in swimming pools), 
city-person to coco-handbag (people carry handbags) and
bdd-fence to ade-cradle (cradles often have a safety fence).

\begin{table*}[htb]
  \centering
  \begin{tabular}{lccccccccc}
    Taxonomy & \# & evals
      & ADE & BDD & City 
      & COCO & IDD & SUN & Vistas \\
    \toprule
    naive concat. & 469 & N/A
    & 27.0 & 55.6 
      & 69.0 & 29.8 & 51.3  
      & 37.4 & 33.7 
    \\
    manual univ.  \cite{Bevandic_2022_WACV} & 294 & N/A
    & 31.0 & 58.5 
      & 72.6 & 35.4 & 54.4 
      & 41.7 & 39.1
    \\
     MSeg original & 194 & N/A & 23.3 & 59.4 
      & 72.6 & 30.3 & 42.6 & 40.2 & 26.1 \\
    \midrule
     auto univ.  &  243 & 164
    & 30.7 & 59.6
      & 72.7 & 35.6 &  55.2
      & 42.3  &  35.8
    \\
    \mytabsep
\end{tabular}
\caption{Multi-domain performance evaluation (mIoU \%) on the MSeg
collection \cite{lambert20cvpr}. Unlike \cite{lambert20cvpr},
we perform evaluation on all 
classes from the original 
dataset taxonomies as in \cite{Bevandic_2022_WACV}.
 }
 \label{table:univ-mseg}
\end{table*}

\section{Conclusion}
We have presented a proof-of-concept
for automatic construction
of interpretable universal
taxonomies for collections of
multi-domain datasets
with incompatible taxonomies.
Our method constructs a set of 1:N mappings
which associate dataset-specific classes with their
universal counterparts.
These mappings establish
a hierarchy of visual concepts
across particular taxonomies
and equip our universal models with
a degree of interpretability. 
The resulting universal taxonomies
allow training in the universal label space 
by treating dataset-specific classes as partial labels.

Our construction approach proceeds
by iterative pairwise
unification.
The unification procedure operates
by testing hypothesized relationships
between dataset specific classes.
We create hypotheses by 
analyzing a bipartite graph between
intra-domain (pseudo) labels 
and extra-domain predictions.
We disambiguate hypotheses 
according to mIoU performance 
of the naive concatenation model 
with post-inference mapping 
on all involved training datasets.

We evaluate our universal taxonomies
in experiments on dataset collections
with incompatible taxonomies.
We consider collections from the same domain 
as well as cross-domain collections.
We use lightweight models
to reduce the training time, 
yet still succeed to infer 
coherent relations between classes.
Our universal models can deliver both 
universal and dataset-specific predictions
without decreasing inference speed.
The reduced number of training logits 
indicates that our models are 
more memory-efficient than ad-hoc alternatives.

Our automatic universal taxonomies
outperform the naive concatenation baseline 
and perform comparably 
to manually designed taxonomies. 
They are also much more flexible 
than custom universal taxonomies 
designed for the standard NLL loss
\cite{lambert20cvpr,Zendel_2022_CVPR}, 
since we can exploit the full training potential
of a given dataset collection without any relabeling effort.
We observe the best relative performance 
of our models in large-scale experiments.

Future work should examine ways of
streamlining the universal taxonomy construction
and explore alternatives for hypothesizing relations
between dataset specific classes. 
\section{Acknowledgement}
    This work 
    has been supported by 
    Croatian Science Foundation
    grant IP-2020-02-5851 ADEPT,
    NVIDIA Academic Hardware Grant Program,
    by European Regional Development Fund
    grant KK.01.1.1.01.0009 DATACROSS
    and by VSITE College for 
    Information Technologies
    who provided access to 6 GPU Tesla-V100 32GB.
    
\bibliographystyle{bmvc2k}
\bibliography{egbib}

\end{document}